\documentclass[a4paper,twoside]{article}

\usepackage{epsfig}
\usepackage{subcaption}
\usepackage{calc}
\usepackage{amssymb}
\usepackage{amstext}
\usepackage{amsmath}
\usepackage{amsthm}
\usepackage{multicol}
\usepackage{pslatex}
\usepackage{apalike}
\usepackage{algorithm2e}
\usepackage{algorithmic}
\usepackage[bottom]{footmisc}
\usepackage{cite}
\usepackage{url}
\usepackage{float}
\usepackage[hidelinks]{hyperref}
\usepackage{eso-pic}
\usepackage{SCITEPRESS}     

\newcommand{\firstpagepublicationfooter}{%
    \AddToShipoutPictureFG*{%
        \AtPageLowerLeft{%
            \raisebox{5mm}[0pt][0pt]{%
                \makebox[\paperwidth][l]{%
                    \hspace*{\dimexpr 1in+\oddsidemargin\relax}%
                    \begin{minipage}[b]{\textwidth}
                        \fontsize{7}{8}\selectfont
                        Zeh, L., Meiwaldt, J., Zhou, Z., Lechler, A. and Verl, A.\\
                        Manipulation of Deformable Linear Objects Using Model Predictive Path Integral Control with Bidirectional Long Short-Term Memory Learning.\\
                        DOI: \href{https://doi.org/10.5220/0013703800003982}{10.5220/0013703800003982}\\
                        Paper published under CC license
                        (\href{https://creativecommons.org/licenses/by-nc-nd/4.0/}{CC BY-NC-ND 4.0})\\
                        In Proceedings of the 22nd International Conference on Informatics in Control, Automation and Robotics (ICINCO 2025) -- Volume 1, pages 47--58\\
                        ISBN: 978-989-758-770-2; ISSN: 2184-2809\\
                        Proceedings Copyright \copyright\ 2025 by SCITEPRESS -- Science and Technology Publications, Lda.
                    \end{minipage}%
                }%
            }%
        }%
    }%
}

\newcommand{\publicationheadertext}{ICINCO 2025 -- 22nd International Conference on Informatics in Control, Automation and Robotics}

\makeatletter
\def\ps@publicationheader{%
    \let\@oddfoot\@empty
    \let\@evenfoot\@empty
    \def\@evenhead{\hfil\slshape\publicationheadertext}%
    \def\@oddhead{{\slshape\publicationheadertext}\hfil}%
    \let\@mkboth\@gobbletwo
    \let\sectionmark\@gobble
    \let\subsectionmark\@gobble
}
\makeatother

\newcommand{\publicationheader}{%
    \setlength{\headheight}{10pt}%
    \addtolength{\headsep}{-10pt}%
    \pagestyle{publicationheader}%
    \thispagestyle{empty}%
}

\begin{document}

\publicationheader
\firstpagepublicationfooter

\title{Manipulation of Deformable Linear Objects Using Model Predictive Path Integral Control with Bidirectional Long Short-Term Memory Learning}

\author{\authorname{Lukas Zeh\sup{1}\orcidAuthor{0000-0003-2730-1383}, Johannes Meiwaldt\sup{1}, Zexu Zhou\sup{1}\orcidAuthor{0009-0002-2163-2528}, Armin Lechler\sup{1}\orcidAuthor{0000-0002-4073-1487} and Alexander Verl\sup{1}\orcidAuthor{0000-0002-2548-6620}}
\affiliation{\sup{1}Institute for Control Engineering of Machine Tools and Manufacturing Units, University of Stuttgart, Stuttgart, Germany}
\email{lukas.zeh@isw.uni-stuttgart.de}
}


\abstract{The manipulation of Deformable Linear Objects (DLOs) such as cables poses a significant challenge for automation due to their infinite degrees of freedom and non-linear dynamics.
In this paper we present a machine learning based optimal control approach for the manipulation of DLOs.
This approach is divided into two main components: modeling and control.
For modeling the dynamics of the DLO, we propose a learning based approach using a bidirectional Long Short-Term Memory (biLSTM) network.
The biLSTM network is trained on synthetic data generated by the MuJoCo physics engine.
For manipulating the DLO, a model predictive control strategy that employs Model Predictive Path Integral (MPPI) control is selected.
The proposed approach is evaluated through simulation and experiments.
The results demonstrate the effectiveness of the proposed method in achieving accurate and efficient manipulation of DLOs.}

\onecolumn \maketitle \normalsize \setcounter{footnote}{0} \vfill

\section{\uppercase{Introduction}}
\label{sec:introduction}

Flexible objects such as textiles, cables or ropes \cite{matsuno_manipulation_2006} can be found almost everywhere, both in everyday life and in the production environment.
They belong to the class of deformable objects \cite{keipour_deformable_2022}.
A sub-category of deformable objects are Deformable Linear Objects (DLOs).
Examples of DLOs include cables, ropes and hoses.
In the context of robotic applications, rigid bodies are typically assumed when gripping and manipulating objects.
This assumption is valid as long as the deformation of the objects is negligible.
However, when handling DLOs, the deformation of the object must be taken into account.
The automated handling of flexible objects by robots is a research problem that has not yet been entirely solved \cite{zhu_challenges_2022, zhou_practical_2020}.

The fundamental challenge in the manipulation of flexible objects, such as DLOs, is that an external force causes both a movement and a change in shape.
Due to the infinite degrees of freedom of DLOs, modeling these nonlinearities during deformation is complex.
Especially for real-time robotic manipulation tasks, accurate and computationally efficient dynamic models are required.
While both physics-based and data-driven approaches exist, each has its own advantages and disadvantages \cite{arriola-rios_modeling_2020}.

To enable effective manipulation of DLOs, Model Predictive Control (MPC) has been successfully employed for planning and control in dynamic environments involving DLOs \cite{yan_self-supervised_2020,wang_offline-online_2022}.
MPC uses a predictive model to simulate and optimize control actions over a finite time horizon, making it suitable for systems with complex, time-varying dynamics.
In the context of DLOs, where deformation must be anticipated and accounted for during manipulation, MPC can utilise a learned or physics-based model to generate feasible, optimized trajectories.

This publication investigates the potential of Model Predictive Path Integral (MPPI) control, a sampling based variant of MPC, for the manipulation of DLOs. 
Simulation data is generated to train the bidirectional Long Short-Term Memory biLSTM network to learn a model of the DLO dynamics offline. 
The model is then used in an MPPI controller to determine the trajectories for manipulating the DLO.

The contribution of our work can be summarized as follows:
\begin{itemize}
    \item{We contribute datasets, model architecture and model weights for modeling cables. The datasets, model architecture and model weights are available at \url{https://doi.org/10.18419/DARUS-5050}.}
    \item{We propose a framework for the manipulation of DLOs that utilizes a Model Predictive Path Integral Controller to manipulate deformable objects.}
    \item{We demonstrate the effectiveness of our proposed method in simulation and experiments.} 
\end{itemize}
The paper is organised as follows: In Section II, we review related work. 
The dataset is introduced in Section III, and our proposed framework is established in Section IV. 
In Section V, we present our simulation and experimental results.

\section{\uppercase{Related Work}}
\label{sec:relatedWork}
The precise manipulation of Deformable Linear Objects (DLOs) requires a physics-based model that accounts for both deformation and an appropriate representation of object shape \cite{sanchez_robotic_2018}.
Modeling approaches can generally be divided into physics-based and data-driven methods \cite{arriola-rios_modeling_2020}.

\subsection{Physics-Based Modeling Approaches}
There are various physics-based modeling approaches for DLOs.
Particle-based models describe DLOs as discrete particles whose positions change in accordance with Newton’s laws under the influence of forces.
In mass-spring-damper systems, these particles are connected by springs, and their physical parameters are described using parameters such as stiffness and damping \cite{schulman_tracking_2013}.
Although these models are computationally efficient, they require precise parameterization, which limits their applicability to real-world industrial cables \cite{monguzzi_optimal_2025}.

Point-based dynamics (PBD), on the other hand, use geometric constraints to directly compute particle positions.
They are more memory- and compute-efficient than mass-spring systems but less physically accurate \cite{arriola-rios_modeling_2020}.

To achieve a more physically accurate representation, the DLO is discretized using Finite Element Methods (FEM) and the deformation equations are solved through numerical integration.
However, FEM approaches are computationally intensive and require accurate material parameters \cite{koessler_efficient_2021, yin_modeling_2021}.
As a result, they are generally unsuitable for real-time robotic manipulation tasks unless specific simplifications are made.

Other numerical methods make assumptions, such as the absence of large deformations, which limits their applicability in more dynamic tasks \cite{rabaetje_real-time_2003}.
Meanwhile, Jacobian-based approaches use local approximations to relate the movement of the robot to the deformation of the object.
While these approaches are real-time capable, they only compute local deformation models \cite{zhu_challenges_2022}.

\subsection{Data-Driven Approaches}
Data-driven models have gained popularity due to their ability to capture the complex nonlinear dynamics of DLOs.
These models are trained using either simulated (offline) data or real-world (online) data.
When using simulated data, physical-based models are typically employed to generate the training data.
The advantage of using simulated data is the ease and speed of data generation compared to collecting real-world data.

Several deep learning approaches have been proposed.
For example, bidirectional Long Short-Term Memory (biLSTM) networks have been used to propagate DLO dynamics over time \cite{yan_self-supervised_2020, yang_learning_2022}.
The interaction-biLSTM proposed by Yang et al. outperformed a baseline biLSTM model in terms of accuracy, although with slightly reduced computational efficiency.

Graph Neural Networks (GNNs) have also been adopted to model DLO dynamics \cite{wang_offline-online_2022, cao_shape_2024}.
In GNN-based methods, the DLO is represented as a graph of discrete capsule elements connected by physically motivated constraints such as bending stiffness, length restrictions, and collisions.
The nodes represent DLO elements, and the edges capture the interactions between them.

Another approach uses radial basis function networks to estimate local deformation models via Jacobian matrices, encoding the relationship between DLO deformation and the robot end-effector position \cite{yu_global_2023}.

While these methods are capable of modeling the complex dynamics of DLOs, they typically require large datasets to achieve robust performance.
It is therefore essential to assess whether models trained on simulation data generalize well enough for real-world robotic manipulation.

\subsection{Model Predictive Control for DLO Manipulation}
Model Predictive Control (MPC) is a strategy used for manipulating DLOs \cite{wang_offline-online_2022}.
It relies on predictive models to simulate object dynamics and optimize control actions over a time horizon.
The predictive model can be either physics-based or learned from data.
MPC is particularly effective for manipulating deformable objects as it enables forward-looking planning that takes into account the evolution of the object's shape.

A sampling-based variant of MPC, Model Predictive Path Integral (MPPI) control, was first introduced in \cite{williams_aggressive_2016} for the autonomous driving of a high-speed RC car. 
In \cite{williams_information_2017}, the authors generalized and formalized the MPPI approach, proposing a learning-based, information-theoretically grounded formulation.
This extension makes MPPI  applicable in data-driven and model-uncertain scenarios.
Since then, MPPI has been applied in various robotic domains \cite{yan_self-supervised_2020, bhardwaj_storm_2021, pezzato_sampling-based_2025}.
The STORM framework, introduced in \cite{bhardwaj_storm_2021} is a fast, sampling based model predictive control framework that works directly in joint space.
It enables real-time responses to complex manipulation tasks, including collisions, joint boundaries and uncertain perception, through GPU parallelization.
\cite{pezzato_sampling-based_2025} use a GPU-based physics simulator as the dynamic model for MPPI control.
This allows high-contact tasks to be solved without explicit modeling or learning, offering a fast, flexible and robust solution in the presence of uncertainties.
\cite{yan_self-supervised_2020} used MPPI control to show the effectiveness of their Coarse-to-fine rope state estimation method.
In their work the MPPI controller is used to estimate the optimal gripping point of the rope for manipulating the rope into a desired shape.
\newline 

We extend existing works by applying MPPI control to manipulate different types of DLOs.
We chose a biLSTM model for modeling the DLO dynamics based on the combination of high inference speed, acceptable accuracy and suitability for robust, flexible control.
To ensure optimal performance, we conduct a hyperparameter search to identify the best model configuration.
The biLSTM model is used in combination with a MPPI controller to generate optimal trajectories for DLO manipulation.
The effectiveness of our approach is demonstrated through simulation and experimental results.

\section{\uppercase{Dataset}}
\label{sec:dataset}
In the following, we describe the dataset used to train the biLSTM model.
\subsection{Simulation Environment}
\begin{figure*}[htbp]
    \centering
    \includegraphics[width=\textwidth]{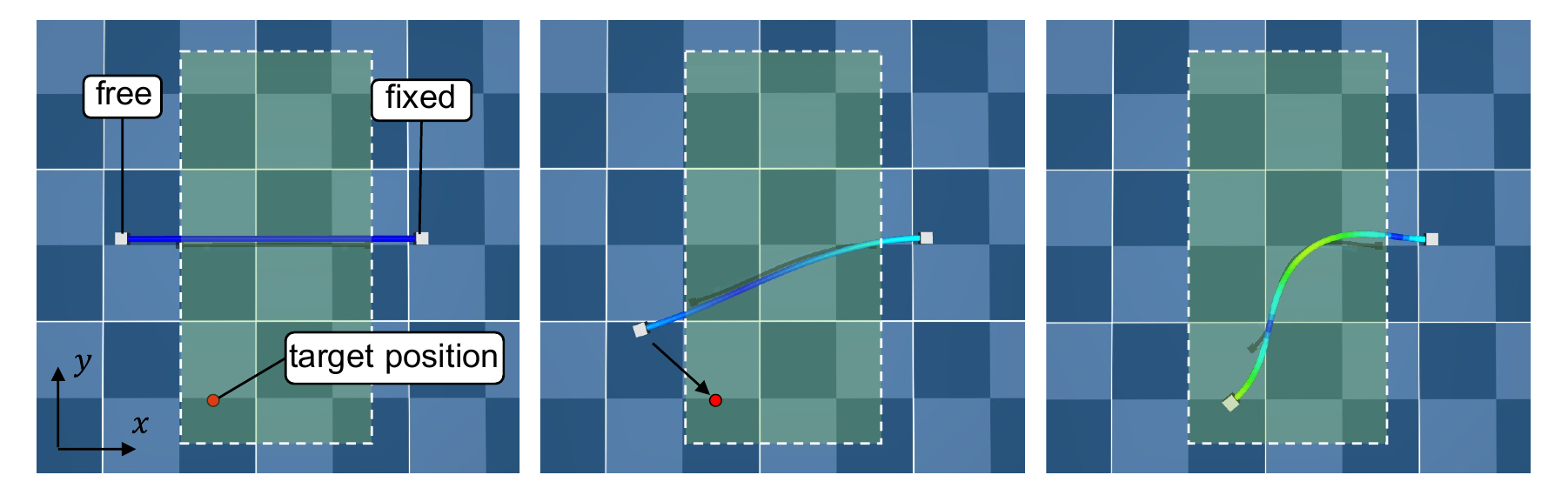}
    \caption{Dataset Generation. The cable is manipulated from a straight line to a random shape by moving the free end of the cable to a random position within the green box.}
    \label{fig:Dataset_Generation}
\end{figure*}

The datasets used to train the biLSTM model were generated using the MuJoCo \cite{todorov_mujoco_2012} physics engine.
MuJoCo natively provides a plugin for the simulation of DLOs.
In this plugin, DLOs are approximated as mass-spring systems.
The DLO is modeled as a chain of mass points, which are connected by linear, torsional, and bending springs.
The individual spring-mass elements are modeled in simplified manner as capsules with the corresponding physical properties.
This approach saves time in the modeling process and also allows for a simple and intuitive implementation.
Furthermore, it has the advantage that the parameters of the DLO can be easily and quickly adjusted, enabling a wide range of DLO variations to be simulated.
To train the biLSTM model, a DLO with a length of 0.5~m was modeled in MuJoCo, consisting of 50 capsules with a diameter of 1~cm.
The number of 50 capsules was chosen as a compromise between realistic behavior and computation time.
The higher the number of capsules, the more degrees of freedom the system has.
This increase in degrees of freedom leads to an almost exponential increase in computation time required to simulate the DLO.
The parameters to be set in the simulation are the Young's modulus [Pa], the shear modulus [Pa], and the damping [Nms/rad] between the individual capsules.
Young's modulus was chosen as $4 \times 10^6$~Pa, the Shear modulus as $1 \times 10^6$~Pa, and the damping was set to 1~Nms/rad.
For the training of the biLSTM model, a simulation step time of 1~ms was chosen.
The individuzal trajectories within the datasets have a length of 5~s.
Influences of gravity, friction, and air drag are not considered in the simulation.
Figure \ref{fig:Dataset_Generation} shows the data generation process.
For the simulation, the DLO is fixed at the right end with a welding condition, so that this end behaves like a clamped end.
The left end of the DLO is manipulated by a robot arm.
The robot arm performs a random trajectory in the xy-plane at a height of 0.15 m.
For the data generation, the cable is manipulated from a straight line into a random shape by moving the left end of the DLO to a random position within the green box.
The target position of the robot is chosen randomly for each trajectory within the range of $\text{x} \in [0.05, 0.35]$~m and $\text{y} \in [-0.2, 0.2]$~m (green area in Figure \ref{fig:Dataset_Generation}).
The origin coordinate system is located in the center of the manipulated capsule.
To generate a wider range of deformations, the DLO is also randomly rotated around the z-axis in the range of $\psi \in [-1, 1]$ rad.
The target position range was chosen to avoid overstretching the DLO.
During the data generation, the positions of the 50 capsules $\mathbf{X}_{DLO}$, the end effector position $\mathbf{X}_{TCP}$, and velocity $\mathbf{v}_{TCP}$ are recorded.

\subsection{Representation of DLO in 2D} 
Since the manipulation takes place on a surface, we chose a representation of the DLO in 2D, similar to \cite{yan_self-supervised_2020}.
For the training of the biLSTM model, the simulation data is reduced.
Instead of using all 50 simulated capsules, in order to reduce the computing time, only every 5th capsule is used for training, resulting in a total of $n = 10$ capsules.
The first and last capsules are also removed, as these are not needed for predicting the DLO dynamics.
The position of the first capsule is described by the pose of the end effector $\mathbf{X}_{TCP}$.
The position of the last capsule remains constant due to the welding condition.
The position of the DLO can therefore be described as a sequence of points in 3D Cartesian space \mbox{$\mathbf{X}_{DLO} \in \mathbb{R}^{n \times 2}$}.
For better generalization of the biLSTM model, the relative position of the capsules with respect to the end effector position $\mathbf{x}_{\text{TCP}}$ is used instead of the absolute position, computed as \mbox{$\mathbf{x}_{r,i} = \mathbf{x}_i - \mathbf{x}_{\text{TCP}}$} for \mbox{$i = 1, \dots, n$}.
For the calculation of the relative positions, only the x and y coordinates are used, as the z-coordinate is constant due to the fixed height of the end effector.
The relative positions of the individual capsules of the DLO are described by
\begin{equation}
    \mathbf{X}_{DLO} = (\mathbf{x}_{r,1}, \mathbf{x}_{r,2}, ..., \mathbf{x}_{r,n}).
\end{equation}
The advantage of this representation is the translational invariance, which allows the neural network to learn the deformation of the DLO not from the absolute positions, but by directly linking the deformation to the end effector position.
The velocity of the capsules is described by the difference of the relative positions at time $t$ and $t-1$.
The overall state of the DLOs is described by
\begin{equation}
    \mathbf{S}_{DLO} = (\mathbf{X}_{DLO}, \dot{\mathbf{X}}_{DLO}).
\end{equation}
The state of the end effector is described by the Cartesian position of the end effector, as well as the rotation of the end effector around the z-axis.
The state of the end effector is therefore described in detail as follows:
\begin{equation}
    \mathbf{S}_{TCP} = (\mathbf{X}_{TCP}, \dot{\mathbf{X}}_{TCP}) = ((x, y, z, \psi), (\dot{x}, \dot{y}, \dot{z}, \dot{\psi})).
\end{equation}
The overall state of the system 
\begin{equation}
    \mathbf{S}_{g} = (\mathbf{S}_{DLO}, \mathbf{S}_{TCP}),
\end{equation}
is obtained by combining the state of the DLO and the state of the end effector.

\section{\uppercase{Proposed Framework}}
\label{sec:proposedFramework}
In this section, we introduce the proposed framework for cable manipulation.
The manipulation of the cable is done in 2D.
First, an overview of the system used for the manipulation task is given.
Then, the bidirectional Long-Short-Term-Memory (biLSTM) model for modeling the DLO dynamics is introduced.
Finally, the Model Predictive Path Integral (MPPI) controller used to manipulate the DLO to the desired shape is described.
\subsection{System Overview}
\begin{figure}[htbp]
    \centerline{\includegraphics[width=0.45\textwidth]{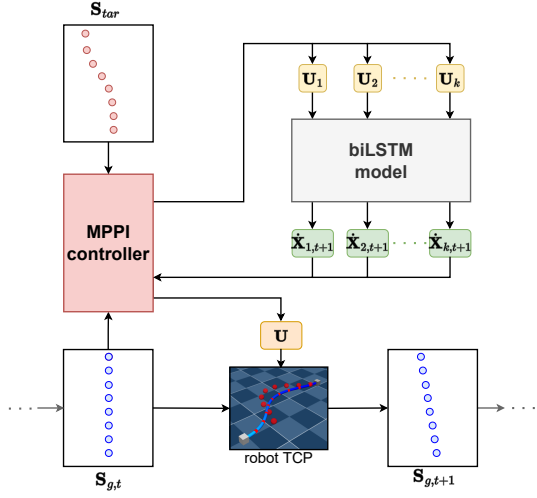}}
    \caption{The proposed framework for cable manipulation uses a biLSTM model trained on synthetic data to predict DLO deformation.
    This prediction is passed to the MPPI controller, which computes the optimal robot control input.}
    \label{fig:MPPI}
\end{figure}

The framework for cable manipulation, as displayed in Figure \ref{fig:MPPI}, consists of a biLSTM model and a MPPI controller.
As an input for the system, the current state of the DLO $\textbf{S}_{g,t}$ and the desired shape of the DLO $\textbf{S}_{tar}$ are used.
The current state of the DLO, $\mathbf{S}_{\text{DLO}}$, is defined by the relative positions and velocities of the capsules, as described in the previous section.
The desired shape of the DLO is represented by the target position of the capsules.
Based on the current state of the DLO and the desired shape, the MPPI controller generates a set of random trajectories $\textbf{U}_i$.
These trajectories are then sent to the biLSTM model, which predicts the velocities of the capsules for the next time step $\dot{\textbf{X}}_i$.
The information about the resulting deformation of the DLO is then used to calculate the cost function for the MPPI controller.
The best trajectory is selected based on the cost function and sent to the robot for execution.
This execution leads to a new state of the DLO $\textbf{S}_{g,t+1}$, which is then used as input for the next iteration of the MPPI controller.
The process is repeated until the DLO has reached the desired shape.

\subsection{biLSTM Cable Model}
\begin{figure}[htbp]
    \centerline{\includegraphics[width=0.45\textwidth]{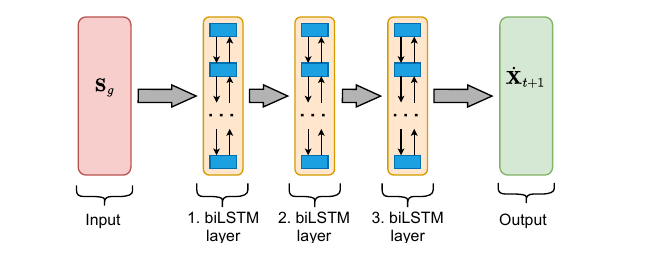}}
    \caption{The biLSTM architecture used for modeling the DLO consists of an input layer, three biLSTM layers and an output layer.}
    \label{fig:biLSTM}
\end{figure}

As in the works of \cite{yang_learning_2022}, \cite{yan_self-supervised_2020}, and \cite{gu_learning_2025}, a biLSTM model is employed for modeling the DLO, as it has been shown to effectively capture its dynamic behavior.
The biLSTM model is a type of recurrent neural network (RNN) that is particularly well-suited to sequence prediction tasks.
The biLSTM is able to capture the dynamics and temporal dependencies of the DLO by processing the sequence of relative positions of the capsules and their velocities.
Unlike standard RNNs, information flows in both temporal directions, allowing the model to use both past and future context for improved sequence understanding.
This feature allows for more effective modeling of relationships along the DLO structure.
This bidirectional processing enhances the LSTM's ability to capture long-range interactions, improving its performance in sequential deformation modeling tasks.
Compared to unidirectional networks such as MLPs, standard RNNs, or unidirectional LSTMs, biLSTMs have advantages in modeling the complex dynamics of deformable linear objects \cite{yang_learning_2022}.
To better capture the coupled dynamics, the biLSTM additionally incorporates the end-effector state as input, enabling the model to learn the interaction between actuator motion and DLO deformation.
Thus, the complete system state $\mathbf{S}_g$ is provided as input to the biLSTM model.
The general structure of the biLSTM model is shown in Figure \ref{fig:biLSTM}.
Its architecture consists of an input layer, one or more stacked biLSTM layers, and a fully connected output layer.
This output layer predicts the capsule velocities for the next time step.
The predicted velocities are then used to compute the cost function for the MPPI controller.

\begin{figure*}[t!]
    \centering
    \includegraphics[width=\textwidth]{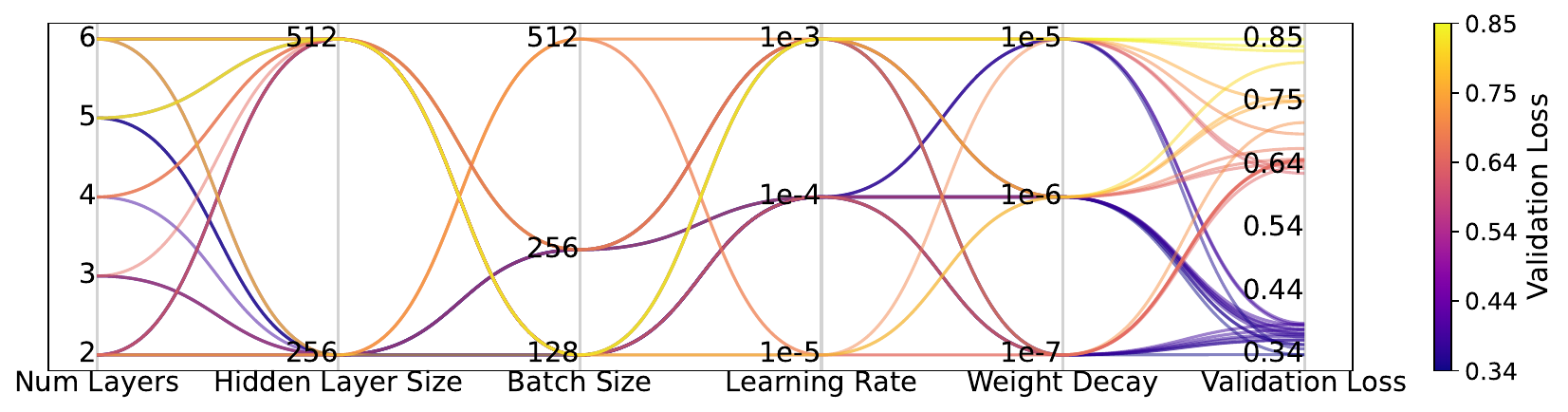}
    \caption{Performance of the biLSTM model in terms of validation loss for different hyperparameter combinations. 
    Visualized are the top 10~\% and the bottom 10~\% of hyperparameter combinations during grid search.}
    \label{fig:hyperparams}
\end{figure*}

\subsubsection{Training}
The biLSTM model is trained on the data generated in the simulation environment.
The biLSTM model is trained using 10,000 trajectories.
These trajectories are split into training and test data, with 80~\% of the data used for training and 20~\% for testing.
The training is performed using the Adam optimizer and the Mean Squared Error (MSE) loss function.
The MSE loss function is defined as:
\begin{equation}
    \text{MSE}(y, \hat{y}) = \frac{1}{N} \sum_{i=1}^{N} (\mathbf{y}_{i} - \hat{\mathbf{y}}_{i})^2,
\end{equation}
where $N$ is the number of nodes, $\mathbf{y}_{i}$ is the true velocity of the node $i$ and $\hat{\mathbf{y}}_{i}$ is the predicted velocity of the node $i$.
To determine the optimal hyperparameters for the biLSTM model, both a random search and a grid search were performed.
The random search was used to perform an initial narrowing down of the hyperparameters.
The random search showed that the most effective models consistently used a hidden layer size of 256 or 512, were trained for up to 100 epochs, and employed a learning rate between 1e-5 and 1e-3.
A low weight decay between 1e-7 and 1e-5 was also common among top-performing configurations.
The number of biLSTM layers varied between 2 and 6, indicating that model depth was less critical compared to other parameters.
Training the model for more than 50 epochs didn't yield significant improvements, suggesting that the model converged well within this range.
In contrast, poor performance was associated with smaller hidden layer sizes, overly small learning rates, high weight decay values, and very small batch sizes.
These results suggest that model capacity, sufficient training duration, and a well-tuned optimization setup are essential for achieving high prediction accuracy.
Based on these findings, a subsequent grid search was then used to find the optimal hyperparameters in a smaller range.
In the table \ref{tab:hyperparameters}, the hyperparameters of the biLSTM model are summarized.
\begin{table}[t]
    \centering
    \caption{Hyperparameters of the biLSTM model, bold values are also used for grid search.}
    \label{tab:hyperparameters}
    \begin{tabular}{|p{0.35\linewidth}|p{0.5\linewidth}|}
        \hline
        \textbf{Hyperparameter} & \textbf{Values Random Search} \\
        \hline
        biLSTM Layers           & [1, \textbf{2, 3, 4, 5, 6}] \\ 
        Hidden Layer Size       & [8, 16, 32, 64, 128, \textbf{256, 512}, 1024] \\
        Epochs                  & [10, 20, 30, 40, \textbf{50}, 100] \\
        Batch Size              & [16, 32, 64, \textbf{128, 256, 512}] \\
        Learning Rate           & [\textbf{1e-3, 1e-4, 1e-5}, 1e-6] \\
        Weight Decay            & [1e-4, \textbf{1e-5, 1e-6, 1e-7}] \\
        \hline
    \end{tabular}
\end{table}

Figure \ref{fig:hyperparams} shows the model performance of various hyperparameter combinations, obtained through grid search, in terms of the validation loss.

Based on the hyperparameter study, the model with the best performance in terms of validation loss was selected.
The model was trained using a learning rate of 1e-4 in combination with a weight decay of 1e-7.
The model with the best performance was trained with batchsize 128 and consists of three biLSTM layers, as shown in Figure \ref{fig:biLSTM}.
Each biLSTM layer consists of 512 neurons (this is equivalent to a hidden layer size of 256 neurons in each direction).
In the following section, the performance of this model is evaluated.

\subsubsection{Model Evaluation}
\begin{figure}[H]
    \centering
    \includegraphics[width=0.45\textwidth]{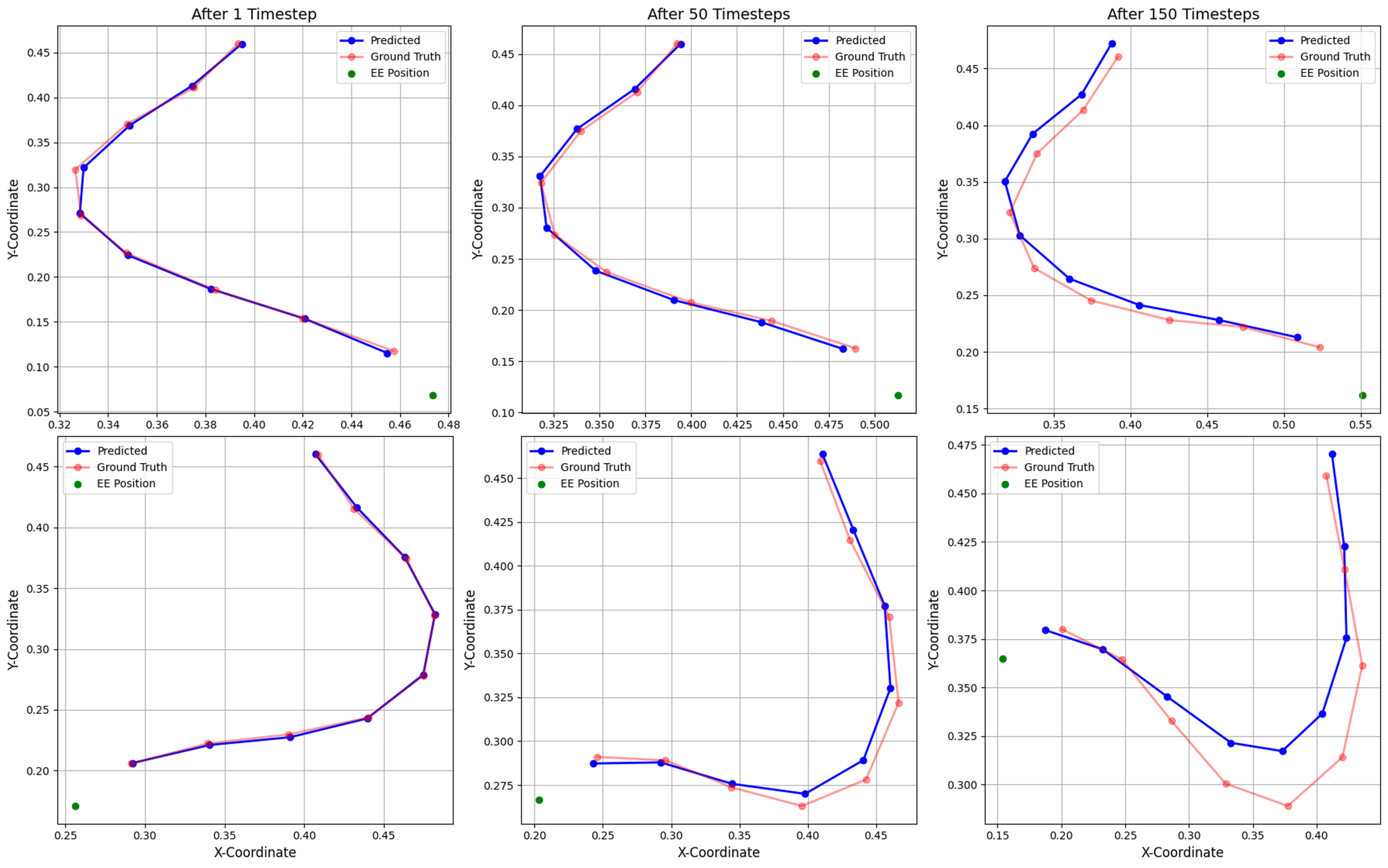}
    \caption{Shape-error $e_\mathrm{shape, biLSTM}$ after 1, 50 and 150 timesteps. The blue line represents the model prediction, while the transparent red line represents the ground truth.}
    \label{fig:Rollout}
\end{figure}

In order to use the biLSTM model in simulation or Model Predictive Control (MPC), a precise rollout prediction over multiple time steps is crucial.
A rollout is a sequence of predicted states over a certain time horizon, which is used to evaluate the model's performance in predicting the DLO dynamics. 
The quality of the dynamic model significantly influences the selection of optimal control sequences.
The model quality is evaluated based on rollouts over 50 time steps (equivalent to 1 second) and 150 time steps (equivalent to 3 seconds).
As a metric for the model quality, the average shape error $e_{\text{shape}}$ and the average velocity error $e_{\text{vel}}$ are used.
\begin{equation}
    e_{\text{shape, biLSTM}} = \|\mathbf{x}_{\text{groundtruth}} - \mathbf{x}_{\text{pred}}\|_2, \quad
\end{equation}
\begin{equation}
    e_{\text{vel, biLSTM}} = \frac{\|\dot{\mathbf{x}}_{\text{groundtruth}} - \dot{\mathbf{x}}_{\text{pred}}\|_2}{\|\dot{\mathbf{x}}_{\text{groundtruth}}\|_2} \times 100~\%.
\end{equation}

The model is evaluated on 100 rollouts, each with a length of 150 time steps (3 seconds).
The average shape error $e_{\text{shape, biLSTM}}$ and the average velocity error $e_{\text{vel, biLSTM}}$ are calculated over all rollouts.
The model is able to predict the shape of the DLO with an average shape error $e_{\text{shape, biLSTM}}$ of 3.3~cm and the velocity with an average velocity error $e_{\text{shape, biLSTM}}$ of 61.59~\% (similar to those in \cite{yang_learning_2022}).
The error of both the shape and the velocity increases with the number of time steps.
Figure \ref{fig:Rollout} shows qualitative results of the shape error over 1, 50 and 150 time steps.
The blue line represents the model prediction, while the transparent red line represents the ground truth.
The biLSTM model shows stable predictions, even over long time intervals.
Structure, length and curvature are preserved, indicating a high model capacity and robust dynamic capture.

\subsection{Model Predictive Path Integral for Manipulation}
Model Predictive Control (MPC) is a well-established control strategy that has been successfully used to manipulate deformable objects \cite{wang_offline-online_2022, yu_shape_2022}.
In this paper, the DLO is manipulated using a Model Predictive Path Integral (MPPI) based control strategy, similar to that in \cite{yan_self-supervised_2020} and \cite{williams_aggressive_2016}.
MPPI is a sampling-based Model Predictive Control strategy particularly suited to handling complex systems and multiple objectives.

In this context, MPPI is employed to compute a control strategy that transforms an initial configuration into a desired target configuration through shape control. 
Actions are represented as target positions for the end effector (EE).

The MPPI algorithm is based on the principle of sampling multiple control sequences around a nominal sequence.
A new control sequence is then generated as a weighted average of these control sequences.
This new sequence is then used to construct the nominal control sequence for the next iteration.

A special feature of MPPI lies in the evaluation of the simulated trajectories.
Each trajectory is assigned a cost value indicating how well the system performs under the respective control inputs.
After simulating numerous future trajectories, each trajectory is assigned a unique set of disturbance values.
The MPPI algorithm calculates a weighted sum of these disturbances.
This nominal control sequence is initialized using one of two alternative mechanisms.
When no prior solution exists, a zero-valued sequence spanning the planning horizon is used as the starting point.
However, when a previous solution is available, a receding horizon approach is employed. 
In this approach, the prior solution is propagated forward by one timestep and the terminal control action is reset to zero. 
This warm-start methodology maintains solution continuity while adhering to the principles of Model Predictive Control.

Each trajectory is evaluated based on its cost value, with lower-cost trajectories receiving higher weights and thus have a greater influence on the control update. 
Specifically, the weight of a trajectory is determined by the exponential function of the negative ratio of its cost to a fixed parameter $\lambda$, also known as temperature~\cite{williams_information_2017}:
\begin{equation}
w_k = e^{-\frac{s_k}{\lambda}},
\end{equation}
where $s_k$ represents the cost of trajectory $k$.
The cost function used in MPPI consists of two terms: a shape cost and a control cost. 
The shape cost penalizes deviations from the target configuration and is defined using the Euclidean seminorm:
\begin{equation}
C_S = 0.5 \cdot \sum_{i=1}^{N} \left( e_{shape, i}^\top \cdot Q \cdot e_{shape, i} \right),
\end{equation}
where $Q = \text{diag}(w_1, w_2, \dots, w_i)$ is a diagonal matrix assigning positive weights $w_i$ to each feature point in the plane. 
The control cost penalizes excessive input effort and is given by:
\begin{equation}
C_R = 0.5 \cdot \sum_{i=1}^{N} \left( u_{b,i}^\top \cdot R \cdot u_{b,i} \right),
\end{equation}
where $R$ is the weighting matrix for the sampled control inputs $u_b$. 

To compute a valid probability distribution over trajectories, the raw weights are normalized:
\begin{equation}
\tilde{w}_k = \frac{w_k}{\sum_{j=1}^{k} w_j}.
\end{equation}
This normalization ensures that trajectories with lower costs contribute more strongly, while keeping the overall influence balanced. 

The MPPI (Model Predictive Path Integral) algorithm proceeds as described in Algorithm~\ref{alg: MPPI}.
It begins with the initialization of a nominal control sequence $\mathbf{U} = \{u_0, u_1, \dots, u_{N-1}\}$, which is typically initialized to zero. 
In each iteration, a set of $K$ trajectories is generated by sampling random disturbances $\delta u_k$ for every time step across the prediction horizon. 
These disturbances are added to the nominal control sequence to create perturbed control sequences.
These are then used in a Monte Carlo tree search.
For generating the random disturbances $\delta u_k$, pink noise \cite{eberhard-2023-pink} is used. 
Each control sequences simulates the system's response to the disturbed input sequence. 
For deformable linear object (DLO) manipulation, this simulation is performed using the biLSTM model, which predicts the resulting DLO states $\dot{\mathbf{X}}_i$ based on the current system state $\mathbf{S}_{g,t}$ and the sampled control input. 
The resulting trajectory is evaluated using a cost function that measures the deviation from the target state $\mathbf{S}_{tar}$ as well as the control effort. 
The total cost of each trajectory $s_k$ is computed, and the corresponding weight $\tilde{w}_k$ is derived as described above. 
The nominal control sequence is then updated using a cost-weighted average of the disturbances:
\begin{equation}
\mathbf{U} \leftarrow \mathbf{U} + \sum_{k=1}^{K} \tilde{w}_k \cdot \delta u_k.
\end{equation}
This update shifts the nominal inputs towards those associated with lower-cost trajectories, thereby gradually improving control performance. 
The updated control input is applied to the system, which advances by one time step.
The resulting new system state is recorded and used as input to the biLSTM model for the next iteration. 
The first element $u_0$ of the optimized control vector $\textbf{U}$, produced by the MPPI controller, is applied to the robot or EE.
The resulting system state is then updated.
This process is repeated until a termination condition is met, either after a fixed number of iterations or when the distance error threshold is reached. 
\begin{algorithm}[h]
    \caption{MPPI Monte-Carlo-Algorithmus.}
    \label{alg: MPPI}
    \KwData{Initial state $\mathbf{S}(g,t_0)$, model dynamics, cost function, prediction horizon $N$, number of control sequences $K$}
    \KwResult{Optimized control input sequence $\mathbf{U}_{0..N-1}$}
    
    initialize control sequence $\mathbf{U}_{0..N-1}$\;
    
    \While{target not reached}{
        generate random disturbances $\delta \mathbf{U}$\;
        \For{control sequences $k = 1..K$}{
            start at current state $\mathbf{X}_{k,0} = \mathbf{X}(t_0)$\;
            \For{horizon steps $n = 0..N-1$}{
                input $\mathbf{U}_{k,n} = \mathbf{U}_n + \delta \mathbf{U}_{k,n}$\;
                next state $\mathbf{X}_{k,n+1} = \text{biLSTM}(\mathbf{X}_{k,n}, \mathbf{U}_{k,n})$\;
                trajectory cost $s_k$ = control cost $C_R$ + shape cost $C_S$\;
            }
        }
        \For{$n = 0..N-1$}{
            $\mathbf{U}_n +=$ reward-weighted disturbance\;
        }
        apply first input $\mathbf{U}_0$ as control input\;
        receive current state\;
        check if target is reached\;
    }
\end{algorithm}
    
    
As shown in Figure~\ref{fig:MPPI}, this complete process enables model-based manipulation of deformable objects by optimizing a control sequence that minimizes cost while adapting to the predicted system dynamics.

\section{\uppercase{Simulation and Experiments}}
In this section, the simulation and experimental results are presented and discussed.
The goal of the simulation and experiments is to analyze the behavior of the DLO and the performance of the MPPI controller.
The simulation and experiments are performed with a Franka Emika Panda robot.
First, the simulation and experimental setup is described.
Then, the results are presented and discussed.

\label{sec:simAndExp}
\subsection{Simulation and Experimental Setup}
\begin{enumerate}
    \item{\textbf{Simulation Setup:}}
    The simulation environment is built using the MuJoCo physics engine.
    As in the simulation, the cable has a length of $50$~cm and a diameter of $1$~cm.
    Young's modulus is $4 \times 10^6$~Pa, the Shear modulus is $1 \times 10^6$~Pa and damping is set to 1~Nms/rad.
    One Franka Emika Panda robot moves one end of the cable so that the shape of the cable matches the desired shape.
    The other end of the cable is fixed.    
    \item{\textbf{Experiment Setup:}}
    The experimental setup is shown in Figure \ref{fig:experiment_setup}.
    A Franka Emika Panda robot is used to manipulate the cable so that the shape of the cable matches the desired shape.
    The other end of the cable is fixed using two zip ties.
    An Intel Realsense D$435$i RGB-D camera is used to track the shape of the cable.
    The biLSTM model and the MPPI controller are implemented on a Ubuntu 24.04 real-time desktop computer.
    The robot trajectories are sent to the robot for execution with a communication frequency of 1,000 Hz.
    The camera data is processed with $40$ fps.
    \begin{figure}
        \centering
        \includegraphics[width=0.45\textwidth]{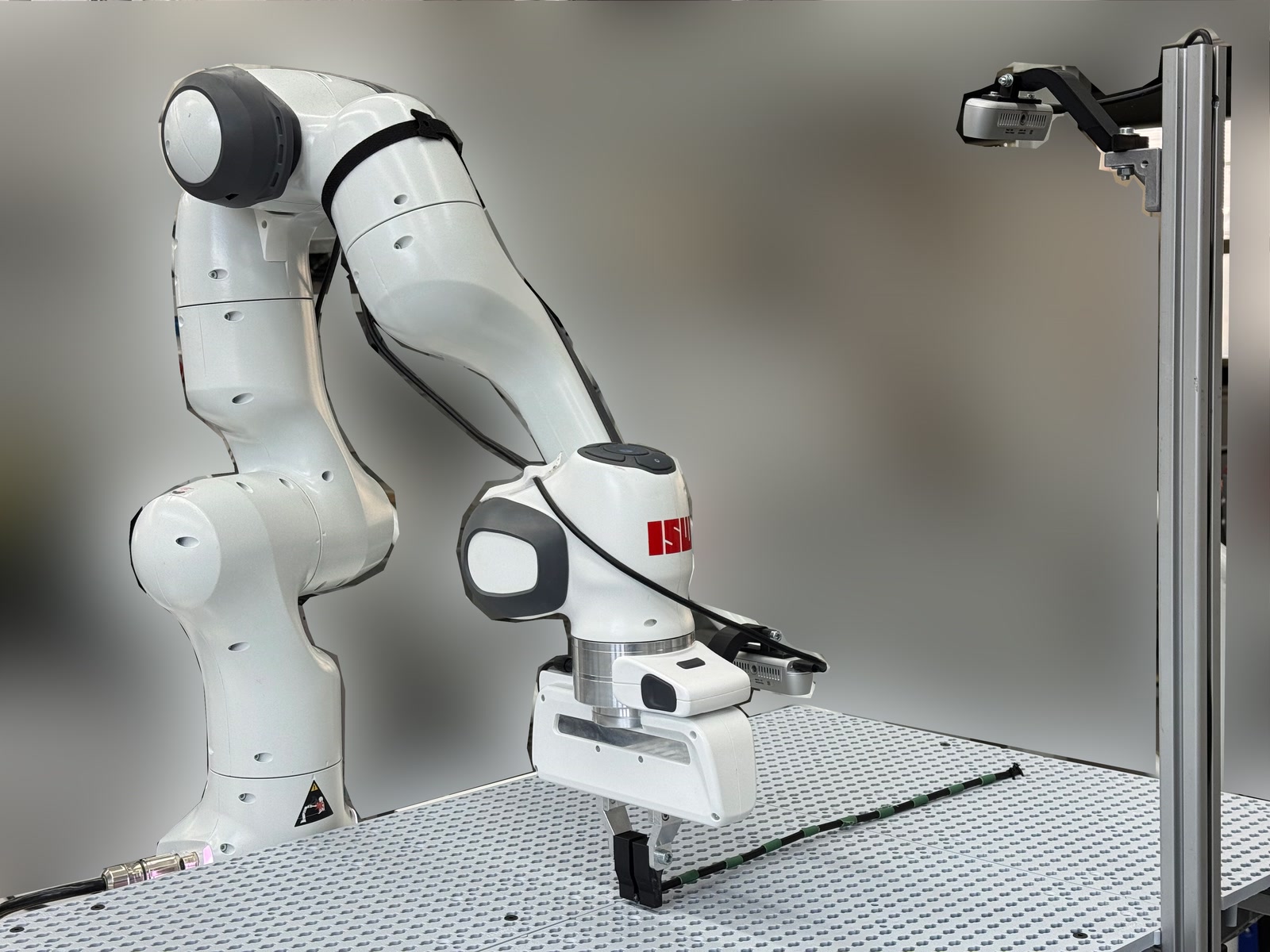}
        \caption{Experiment Setup.}
        \label{fig:experiment_setup}
    \end{figure}
\end{enumerate}

\subsection{Simulation Results}
\begin{figure}
    \centering
    \includegraphics[width=0.45\textwidth]{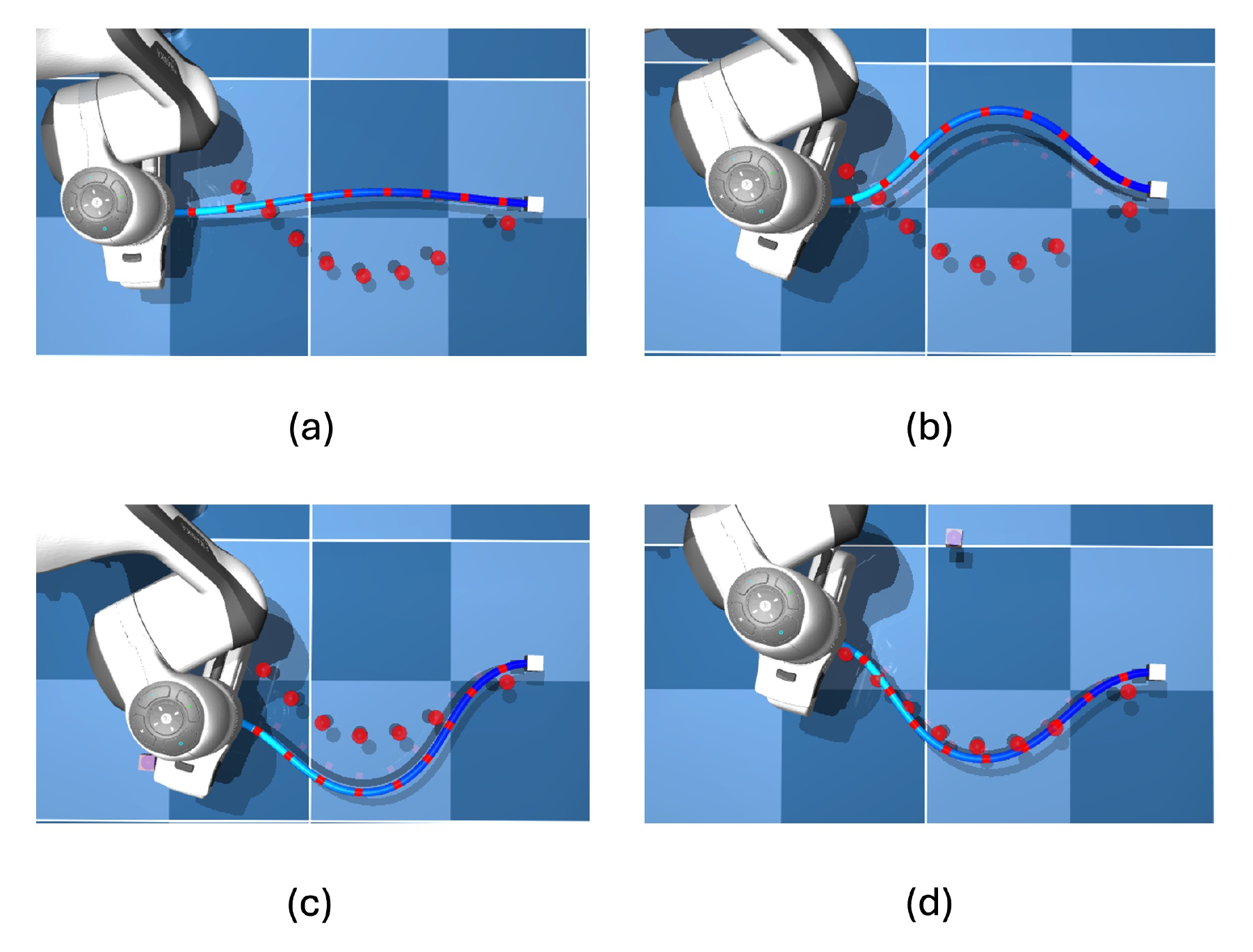}
    \caption{Steps of the shape control in the simulation environment.}
    \label{fig:U_shape}
\end{figure}

In the following, the results of the simulation are presented.
The simulation was performed using the MuJoCo model of the Franka Emika Panda robot provided by the MuJoCo physics engine.
The left end of the DLO is firmly gripped by the end effector of the robot.
The right end of the DLO is fixed.
The initial pose of the robot is set to reflect the initial position of the DLO, meaning the robot is positioned so that the DLO is in a straight line.
Each control output of the MPPI controller is set as the target position for a motion capture body.
This body serves as the Cartesian target position for the inverse kinematics control of the robot.
The control process was performed in the x-y-plane, while the z-position was kept constant at 10~cm.
The rotation around the z-axis also received a direct control input from the MPPI output.
The control values were limited to a range of -0.2~m to 0.2~m in the x-direction and 0~m to 0.35~m in the y-direction.
This reflects the workspace during the generation of the training data.
The parameters of the MPPI controller are displayed in Table \ref{tab:MPPI_params_sim}.
These parameters were determined empirically through a series of experiments.
\begin{table}[htbp]    
        \centering
        \caption{MPPI parameters in simulation.}
        \label{tab:MPPI_params_sim}
        \begin{tabular}{|p{0.54\linewidth}|p{0.31\linewidth}|}
            \hline
            \textbf{Parameter} & \textbf{Value} \\
            \hline
            Horizon ($H$)                                       & 5 \\ 
            Time increment ($dt$)                               & 0,2 \\
            Number of samples ($N$)                             & 20 \\
            Temperature ($\lambda$)                             & 0.002 \\
            Standard deviation of the disturbances ($\delta$)   & $[0.5; 0.5; 0; 3]$ \\
            for $[X, Y, Z, \text{Rot}(Z)]$                      & \\
            Form error weighting ($Q$)                      & 150 \\
            Control error weighting ($R$)                   & 5 \\
            \hline
        \end{tabular}
\end{table}
First, the performance of the MPPI controller is evaluated by shaping the DLO into a U-shape.
Figure \ref{fig:U_shape} shows the process of the shape control in the simulation.
In (a), the initial state of the DLO is shown.
In (b), the DLO is deformed to the opposite side of the target position. 
In (c), the MPPI control could compensate for the initial deformation and move the DLO to the other side. 
In (d), the MPPI control has approached the target position and the trial has ended.
100 trials were performed, with a success criterion being a maximum deviation of 2~cm between the positions of the simulated capsules (represented by red dots on the cable) and their corresponding target points (represented by red dots on the plane).
The success rate was 93~\%, with an average time of 7.26~s per successful trial.
A successful trial is defined as a trial in which the DLO is shaped into the desired U-shape within 30 seconds.

Additionally, the performance of the MPPI controller is evaluated over 1,000 trials of shaping the DLO into random goal shapes.
The success rate was 20.5~\%, with an average time of 13.3~s per successful trial.

The simulation results show that the MPPI-based control in combination with a biLSTM dynamics model is able to shape the DLO into a desired target shape.
The approach combines sample-based control with a neural network for modeling the DLO dynamics.
By training with relative positions of the capsules, the approach can also be transferred to new scenarios.
However, re-optimizing the MPPI parameters is necessary in scenarios where the target shape differs significantly from the one for which the controller was originally tuned.
Additional parameterization is also required to handle severe deformations of the DLO effectively.
One limiting factor is the computational demand of the biLSTM model, especially when processing a large number of fault samples. Increasing computing resources could improve the controller's performance, as a larger sample size generally leads to greater accuracy and robustness.
Another potential bottleneck is the current sampling strategy used by the MPPI controller.
Using an adaptive sampling approach where exploration during the construction of the search tree focuses only on trajectories with high solution potential, might improve the results.

\subsection{Experiment Results}
The performance of the MPPI controller is evaluated across three different scenarios.
In the first scenario, 2D shape control is performed on a 50~cm long cable with a diameter of 6~mm, equipped with 9 markers.
The second scenario uses the same cable and marker setup, but shape control is conducted in 3D.
The third scenario involves 2D shape control of a 50~cm long wire with a diameter of 1.5~mm, without any markers.
While initial parameter tuning was performed in simulation, further adjustments were necessary during practical validation to compensate for discrepancies between simulated and real-world behavior.
The MPPI controller parameters remain consistent across all scenarios and are listed in Table \ref{tab:MPPI_params_exp}.

\begin{table}[htbp]    
        \centering
        \caption{MPPI parameters in experiment.}
        \label{tab:MPPI_params_exp}
        \begin{tabular}{|p{0.54\linewidth}|p{0.31\linewidth}|}
            \hline
            \textbf{Parameter} & \textbf{Value} \\
            \hline
            Horizon ($H$)                                       & 8 \\ 
            Time increment ($dt$)                               & 0.02 \\
            Number of samples ($N$)                             & 200 \\
            Temperature ($\lambda$)                             & 0.002 \\
            Standard deviation of the disturbances ($\delta$)   & $[0.2; 0.2; 0; 0.02]$ \\
            for $[X, Y, Z, \text{Rot}(Z)]$                      & \\
            Form error weighting ($Q$)                          & 150 \\
            Control error weighting ($R$)                       & 5 \\
            \hline
        \end{tabular}
\end{table}
In all three scenarios, the cable is manipulated into a U-shape.
The goal of the shape control experiments is to shape the cable into the desired U-shape with a maximum deviation of 2~cm from the target position.
The derivation of the target position is based on the difference between the measured positions of the markers on the cable and the target positions of the markers.
In the first and second scenario, 9 markers along the cable are used to track the shape of the cable.
The positions of these markers are tracked using a color filter.
In the third scenario, no markers are used to track the shape of the cable.
Instead, the FastDLO algorithm \cite{caporali_fastdlo_2022} is used for shape estimation.
Based on the estimated shape, 9 virtual markers are placed along the tracked shape in order to keep the process of the shape control as similar as possible to the first two scenarios.
\newline 

\begin{figure}[htbp]
    \centering
    \includegraphics[width=0.45\textwidth]{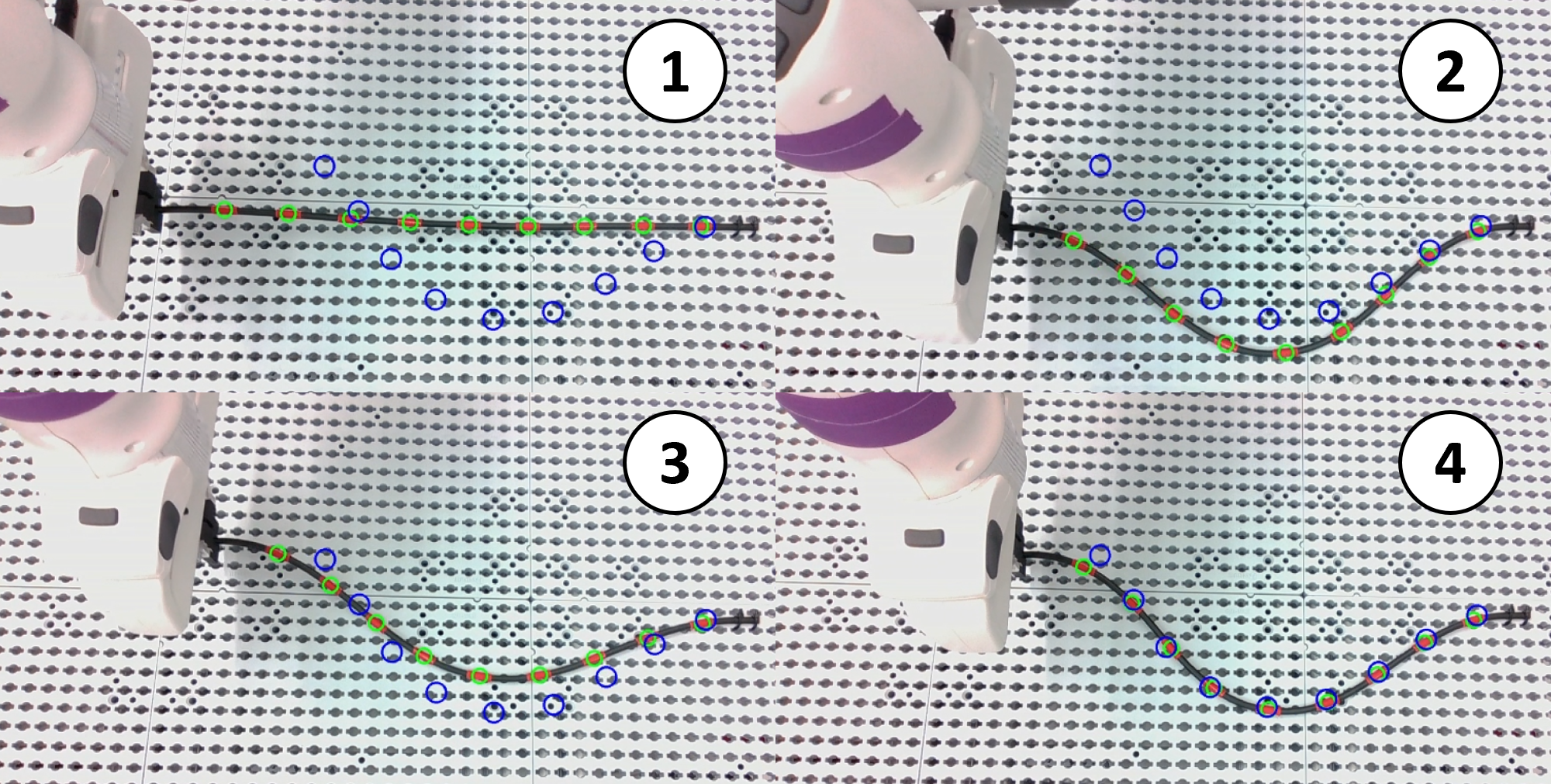}
    \caption{Experiment result of the 2D shape control of a 50~cm long cable with a diameter of 6~mm, marked with 9 markers.
    The cable is manipulated into a U-shape.}
    \label{fig:U_shape_exp}
\end{figure}

\begin{figure}[htbp]
    \centering
    \includegraphics[width=0.45\textwidth]{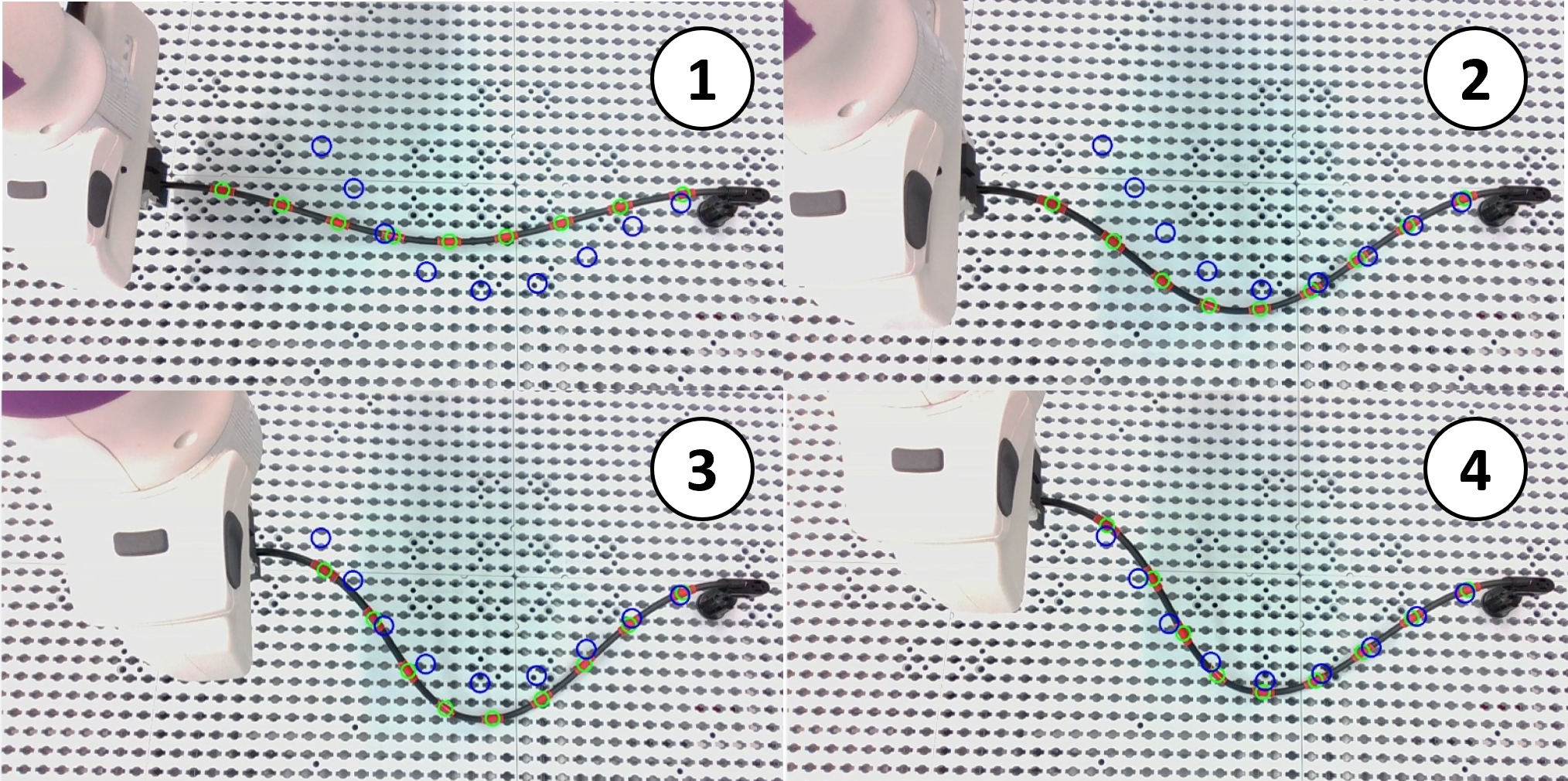}
    \caption{Experiment result of the 3D shape control of a 50~cm long cable with a diameter of 6~mm, marked with 9 markers.
    The cable is manipulated into a U-shape.}
    \label{fig:U_shape_exp_3D}
\end{figure}

\begin{figure}[htbp]
    \centering
    \includegraphics[width=0.45\textwidth]{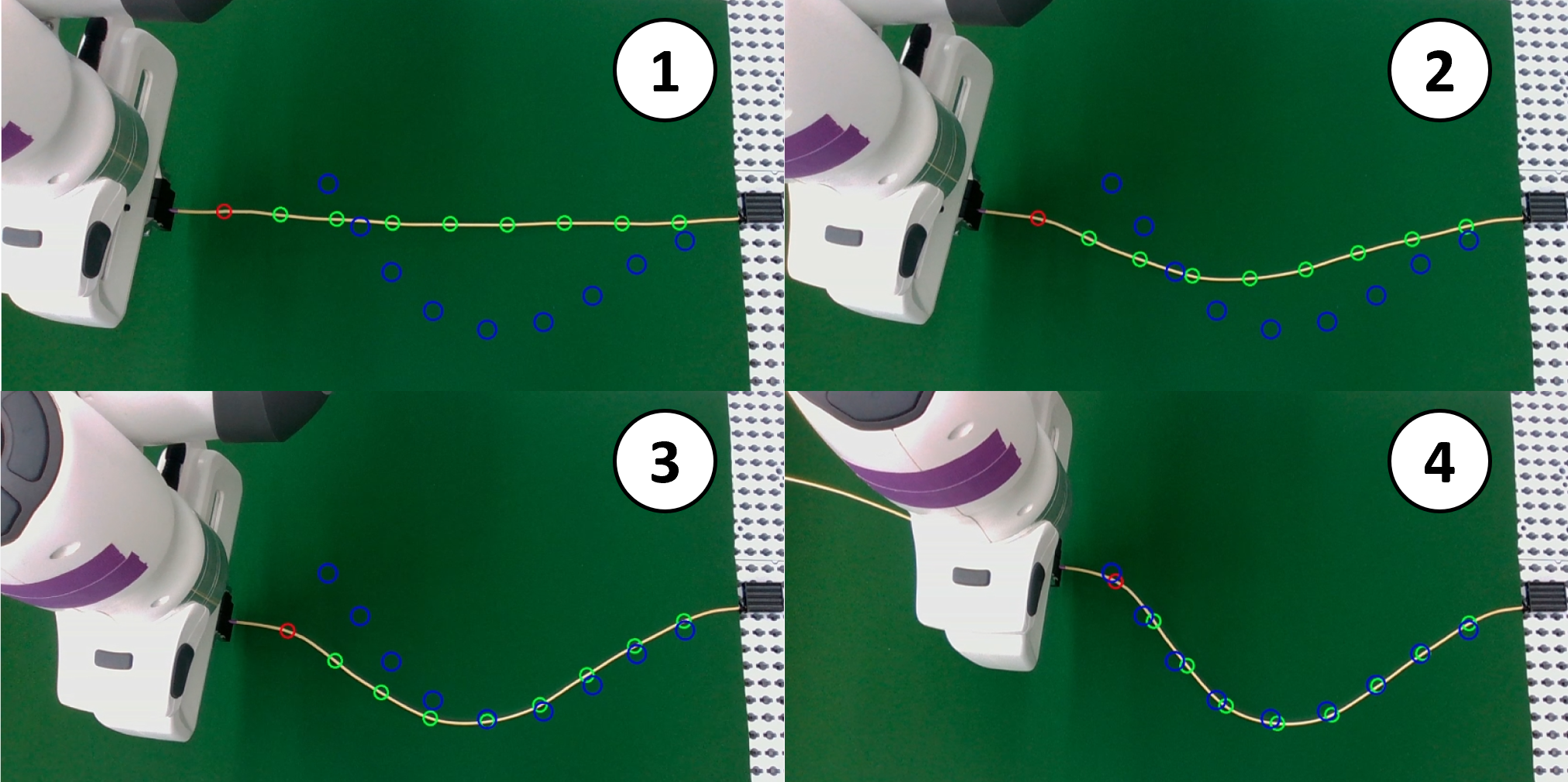}
    \caption{Experiment result of the 2D shape control of a 50~cm long wire with a diameter of 1.5~mm.
    The cable is manipulated into a U-shape.}
    \label{fig:U_shape_exp_FLYR}
\end{figure}

In the first scenario, the cable is mounted directly on the working plane.
In Figure \ref{fig:U_shape_exp}, the process of a successful 2D shape control is shown.
Indicated are the initial state of the cable and robot (1), two intermediate states (2) and (3), and the final state of the cable and robot (4), where the cable has been successfully shaped into the desired U-shape.
To evaluate the performance of the MPPI controller, 20 trials were performed.
The success rate was 85~\%, with an average time of 15.7~s per successful trial.
The fastest trial took 2.1~s, and the slowest took 45.2~s.
\newline
In the second scenario, the cable is mounted on a 7~cm high platform.
As in the first scenario, the 9 markers are used to track the shape of the cable.
The positions of these markers are tracked by using a color filter.
In Figure \ref{fig:U_shape_exp_3D}, the process of a successful 3D shape control is shown.
The starting position of the robot remains the same as in the first scenario.
This leads to the cable being slightly bent by gravity at the beginning ,which can be seen in (1).
In (2) and (3), two intermediate states of the shape control are displayed.
The final state of the cable and robot can be seen in (4), where the cable has been successfully shaped into the desired U-shape.
Even though this scenario was not trained, the MPPI controller is able to manipulate the cable into a U-shape.
The process of the shape control is not as roboust as in the first scenario, leading to a higher failure rate.
In the pracical experiments, failure was defined as a trial in which the DLO is not shaped into the desired U-shape within 90 seconds.
Also, the time to reach the target position is significantly higher.
To evaluate the controller, 20 trials were performed.
The success rate was 50~\%, with an average time of 28.3~s per successful trial.
The fastest trial took 10.4~s, and the slowest 80.5~s.
\newline
In the third scenario, we used a thinner and more flexible wire with a diameter of 1.5~mm.
This wire was selected to test the performance of the biLSTM model in respect to generalization.
Like in the first scenario, the wire is mounted directly on the working plane.
In this scenario, no markers are used to track the shape of the wire, as we want to test the performance in a more real-world-like scenario.
Instead, the FastDLO algorithm \cite{caporali_fastdlo_2022} is used for shape estimation.
Since the background of the testbench is white and the wire used is light yellow, we needed to use a greenscreen to be able to track the shape of the wire.
In Figure \ref{fig:U_shape_exp_FLYR}, the process of a successful 2D shape control is shown.
The starting position of the robot remains the same as in the first scenario.
In (1), the initial state of the wire and Robot is shown.
In (2) and (3), two intermediate states of the shape control are displayed.
In (4), the final state of the wire and robot can be seen, where the wire has been successfully shaped into the desired U-shape.
Like in the second scenario, this scenario is less robust than the first scenario.
Also leading to a higher failure rate and a longer time to reach the target position.
To evaluate the controller, 20 trials were performed.
The success rate was 60~\%, with an average time of 25.4~s per successful trial.
The fastest trial took 8.2~s and the slowest 65.3~s.
\newline

Additionally to the shapes shown in the figures, we also performed trials with different target shapes.
The more the target shapes resemble a U-shape, the more the more likely the controller is to successfully shape the DLO into the desired shape.
As the simulation results have shown, this control approach struggles especially with shapes requiring significant deformations.
Since the cable and wire used in the experiments are pre-bend in one direction, the controller also struggles to deform the DLOs in the direction opposite to the pre-bend.
Additionally, the approach is likely to fail if the DLO is deformed in the wrong direction at the beginning of the trial.
\newline

Besides these issues and limitations, the results of the experiments show that the MPPI-based control in combination with a biLSTM dynamics model, is able to shape the DLO into a desired target shape.
The experiments have also shown that the biLSTM model is able to model different kinds of DLOs with very different properties.
As mentioned before, the MPPI controller has to be retuned for scenarios that are not similar to the scenario for which the controller was initially tuned.

\section{\uppercase{Discussion and Future Works}}
\label{sec:conclusion}
In this paper, we presented a novel approach for the manipulation of deformable linear objects (DLOs) using a biLSTM model and a Model Predictive Path Integral (MPPI) controller.
The approach combines a neural network for modeling the DLO dynamics with a sampling-based control strategy.
The biLSTM model is trained on a dataset of simulated DLO trajectories, which are generated using a MuJoCo model of the DLO.
The model is able to predict the shape and velocity of the DLO over multiple time steps.
The MPPI controller is used to manipulate the DLO into a desired target shape.
The approach was evaluated in simulation and experiments using a Franka Emika Panda robot.
The results show that the MPPI-based control in combination with a biLSTM dynamics model is able to shape the DLO into a desired target shape.
The approach is able to generalize to different kinds of DLOs with different physical properties.

In the future, we want to tune the pretreained models with real data in order to minimize the sim to real gap and obtain more robust models.
We also want to test different kinds of neural networks inside the Model Predictive Control loop, in order to evaluate if the approach becomes more robust or faster when faster or more accurate models are used.
In respect to faster models, we want to test the performance of MLPs.
In respect to more accurate models, we want to test the performance of GNNs.

In order to improve the performance of the MPPI controller, we want to investigate different sampling strategies and different cost functions.
\section*{\uppercase{Acknowledgements}}
This work was funded by Deutsche Forschungsgemeinschaft (DFG, German Research Foundation) under Germany´s Excellence Strategy – EXC 2075 – 390740016.

\bibliographystyle{apalike}
{\small
\bibliography{references}}

@inproceedings{koessler_efficient_2021,
	title = {An efficient approach to closed-loop shape control of deformable objects using finite element models},
	isbn = {978-1-7281-9077-8},
	url = {https://ieeexplore.ieee.org/document/9560919/},
	doi = {10.1109/ICRA48506.2021.9560919},
	language = {en},
	urldate = {2025-02-12},
	booktitle = {2021 {IEEE} {International} {Conference} on {Robotics} and {Automation} ({ICRA})},
	author = {Koessler, A. and Filella, N. Roca and Bouzgarrou, B.C. and Lequievre, L. and Ramon, J.-A. Corrales},
	month = may,
	year = {2021},
}

@inproceedings{yu_shape_2022,
	title = {Shape {Control} of {Deformable} {Linear} {Objects} with {Offline} and {Online} {Learning} of {Local} {Linear} {Deformation} {Models}},
	isbn = {978-1-7281-9681-7},
	url = {https://ieeexplore.ieee.org/document/9812244/},
	doi = {10.1109/ICRA46639.2022.9812244},
	language = {en},
	urldate = {2025-04-24},
	booktitle = {2022 {International} {Conference} on {Robotics} and {Automation} ({ICRA})},
	author = {Yu, Mingrui and Zhong, Hanzhong and Li, Xiang},
	month = may,
	year = {2022},
}

@inproceedings{williams_information_2017,
	title = {Information theoretic {MPC} for model-based reinforcement learning},
	isbn = {978-1-5090-4633-1},
	url = {https://ieeexplore.ieee.org/document/7989202/},
	doi = {10.1109/ICRA.2017.7989202},
	language = {en},
	urldate = {2025-03-19},
	booktitle = {2017 {IEEE} {International} {Conference} on {Robotics} and {Automation} ({ICRA})},
	author = {Williams, Grady and Wagener, Nolan and Goldfain, Brian and Drews, Paul and Rehg, James M. and Boots, Byron and Theodorou, Evangelos A.},
	month = may,
	year = {2017},
}

@inproceedings{williams_aggressive_2016,
	title = {Aggressive driving with model predictive path integral control},
	isbn = {978-1-4673-8026-3},
	url = {https://ieeexplore.ieee.org/document/7487277/},
	doi = {10.1109/ICRA.2016.7487277},
	language = {en},
	urldate = {2025-02-12},
	booktitle = {2016 {IEEE} {International} {Conference} on {Robotics} and {Automation} ({ICRA})},
	author = {Williams, Grady and Drews, Paul and Goldfain, Brian and Rehg, James M. and Theodorou, Evangelos A.},
	month = may,
	year = {2016},
}

@inproceedings{todorov_mujoco_2012,
	title = {{MuJoCo}: {A} physics engine for model-based control},
	isbn = {978-1-4673-1736-8 978-1-4673-1737-5 978-1-4673-1735-1},
	shorttitle = {{MuJoCo}},
	url = {http://ieeexplore.ieee.org/document/6386109/},
	doi = {10.1109/IROS.2012.6386109},
	language = {en},
	urldate = {2025-02-12},
	booktitle = {2012 {IEEE}/{RSJ} {International} {Conference} on {Intelligent} {Robots} and {Systems}},
	author = {Todorov, Emanuel and Erez, Tom and Tassa, Yuval},
	month = oct,
	year = {2012},
}

@inproceedings{schulman_tracking_2013,
	title = {Tracking deformable objects with point clouds},
	isbn = {978-1-4673-5643-5 978-1-4673-5641-1},
	url = {http://ieeexplore.ieee.org/document/6630714/},
	doi = {10.1109/ICRA.2013.6630714},
	language = {en},
	urldate = {2025-02-06},
	booktitle = {2013 {IEEE} {International} {Conference} on {Robotics} and {Automation}},
	author = {Schulman, John and Lee, Alex and Ho, Jonathan and Abbeel, Pieter},
	month = may,
	year = {2013},
}

@inproceedings{rabaetje_real-time_2003,
	title = {Real-time {Simulation} of {Deformable} {Objects} for {Assembly} {Simulations}},
	isbn = {0-909925-96-8},
	doi = {10.5555/820086.820101},
	language = {en},
	booktitle = {Proceedings of the {Fourth} {Australasian} {User} {Interface} {Conference} on {User} {Interfaces} 2003 - {Volume} 18},
	author = {Rabaetje, Ralf},
	year = {2003},
}

@article{matsuno_manipulation_2006,
	title = {Manipulation of deformable linear objects using knot invariants to classify the object condition based on image sensor information},
	volume = {11},
	doi = {10.1109/tmech.2006.878557},
	number = {4},
	journal = {IEEE/ASME Transactions on Mechatronics},
	author = {Matsuno, T. and Tamaki, D. and Arai, F. and Fukuda, T.},
	month = aug,
	year = {2006},
}

@article{zhu_challenges_2022,
	title = {Challenges and {Outlook} in {Robotic} {Manipulation} of {Deformable} {Objects}},
	volume = {29},
	doi = {10.1109/mra.2022.3147415},
	number = {3},
	journal = {IEEE Robotics and Automation Magazine},
	author = {Zhu, Jihong and Cherubini, Andrea and Dune, Claire and Navarro-Alarcon, David and Alambeigi, Farshid and Berenson, Dmitry and Ficuciello, Fanny and Harada, Kensuke and Kober, Jens and Li, Xiang and Pan, Jia and Yuan, Wenzhen and Gienger, Michael},
	month = sep,
	year = {2022},
}

@inproceedings{zhou_practical_2020,
	title = {A {Practical} {Solution} to {Deformable} {Linear} {Object} {Manipulation}: {A} {Case} {Study} on {Cable} {Harness} {Connection}},
	doi = {10.1109/icarm49381.2020.9195380},
	booktitle = {2020 5th {International} {Conference} on {Advanced} {Robotics} and {Mechatronics} ({ICARM})},
	publisher = {IEEE},
	author = {Zhou, Hang and Li, Shunchong and Lu, Qi and Qian, Jinwu},
	month = dec,
	year = {2020},
}

@article{yin_modeling_2021,
	title = {Modeling, learning, perception, and control methods for deformable object manipulation},
	volume = {6},
	issn = {2470-9476},
	url = {https://www.science.org/doi/10.1126/scirobotics.abd8803},
	doi = {10.1126/scirobotics.abd8803},
	language = {en},
	number = {54},
	urldate = {2025-03-03},
	journal = {Science Robotics},
	author = {Yin, Hang and Varava, Anastasia and Kragic, Danica},
	month = may,
	year = {2021},
}

@article{wang_offline-online_2022,
	title = {Offline-{Online} {Learning} of {Deformation} {Model} for {Cable} {Manipulation} with {Graph} {Neural} {Networks}},
	volume = {7},
	issn = {2377-3766, 2377-3774},
	url = {http://arxiv.org/abs/2203.15004},
	doi = {10.1109/LRA.2022.3158376},
	language = {en},
	number = {2},
	urldate = {2025-02-06},
	journal = {IEEE Robotics and Automation Letters},
	author = {Wang, Changhao and Zhang, Yuyou and Zhang, Xiang and Wu, Zheng and Zhu, Xinghao and Jin, Shiyu and Tang, Te and Tomizuka, Masayoshi},
	month = apr,
	year = {2022},
}

@article{keipour_deformable_2022,
	title = {Deformable {One}-{Dimensional} {Object} {Detection} for {Routing} and {Manipulation}},
	volume = {7},
	doi = {10.1109/lra.2022.3146920},
	number = {2},
	journal = {IEEE Robotics and Automation Letters},
	author = {Keipour, Azarakhsh and Bandari, Maryam and Schaal, Stefan},
	month = apr,
	year = {2022},
}

@misc{yan_self-supervised_2020,
	title = {Self-{Supervised} {Learning} of {State} {Estimation} for {Manipulating} {Deformable} {Linear} {Objects}},
	url = {http://arxiv.org/abs/1911.06283},
	doi = {10.48550/arXiv.1911.06283},
	urldate = {2024-08-15},
	publisher = {arXiv},
	author = {Yan, Mengyuan and Zhu, Yilin and Jin, Ning and Bohg, Jeannette},
	month = jun,
	year = {2020},
}

@article{pezzato_sampling-based_2025,
	title = {Sampling-{Based} {Model} {Predictive} {Control} {Leveraging} {Parallelizable} {Physics} {Simulations}},
	volume = {10},
	issn = {2377-3766, 2377-3774},
	url = {https://ieeexplore.ieee.org/document/10856359/},
	doi = {10.1109/LRA.2025.3535185},
	language = {en},
	number = {3},
	urldate = {2025-03-19},
	journal = {IEEE Robotics and Automation Letters},
	author = {Pezzato, Corrado and Salmi, Chadi and Trevisan, Elia and Spahn, Max and Alonso-Mora, Javier and Hernández Corbato, Carlos},
	month = mar,
	year = {2025},
}

@article{arriola-rios_modeling_2020,
	title = {Modeling of {Deformable} {Objects} for {Robotic} {Manipulation}: {A} {Tutorial} and {Review}},
	doi = {10.3389/frobt.2020.00082},
	journal = {Frontiers in Robotics and AI},
	author = {Arriola-Rios, Veronica E. and Guler, Puren and Ficuciello, Fanny and Kragic, Danica and Siciliano, Bruno and Wyatt, Jeremy L.},
	month = sep,
	year = {2020},
}

@misc{bhardwaj_storm_2021,
	title = {{STORM}: {An} {Integrated} {Framework} for {Fast} {Joint}-{Space} {Model}-{Predictive} {Control} for {Reactive} {Manipulation}},
	shorttitle = {{STORM}},
	url = {http://arxiv.org/abs/2104.13542},
	doi = {10.48550/arXiv.2104.13542},
	language = {en},
	urldate = {2025-04-01},
	publisher = {arXiv},
	author = {Bhardwaj, Mohak and Sundaralingam, Balakumar and Mousavian, Arsalan and Ratliff, Nathan and Fox, Dieter and Ramos, Fabio and Boots, Byron},
	month = sep,
	year = {2021},
}

@article{cao_shape_2024,
	title = {Shape {Control} of {Elastic} {Deformable} {Linear} {Objects} for {Robotic} {Cable} {Assembly}},
	volume = {6},
	issn = {2640-4567, 2640-4567},
	url = {https://onlinelibrary.wiley.com/doi/10.1002/aisy.202300835},
	doi = {10.1002/aisy.202300835},
	language = {en},
	number = {7},
	urldate = {2025-02-06},
	journal = {Advanced Intelligent Systems},
	author = {Cao, Bin and Zang, Xizhe and Zhang, Xuehe and Chen, Zhuo and Li, Shouqiang and Zhao, Jie},
	month = jul,
	year = {2024},
}

@article{caporali_fastdlo_2022,
	title = {{FASTDLO}: {Fast} {Deformable} {Linear} {Objects} {Instance} {Segmentation}},
	volume = {7},
	url = {https://github.com/lar-unibo/fastdlo},
	doi = {10.1109/lra.2022.3189791},
	number = {4},
	journal = {IEEE Robotics and Automation Letters},
	author = {Caporali, Alessio and Galassi, Kevin and Zanella, Riccardo and Palli, Gianluca},
	month = oct,
	year = {2022},
}

@article{gu_learning_2025,
	title = {Learning {Graph} {Dynamics} with {Interaction} {Effects} {Propagation} for {Deformable} {Linear} {Objects} {Shape} {Control}},
	issn = {1558-3783},
	url = {https://ieeexplore.ieee.org/document/10845081},
	doi = {10.1109/TASE.2025.3530957},
	urldate = {2025-03-27},
	journal = {IEEE Transactions on Automation Science and Engineering},
	author = {Gu, Feida and Sang, Hongrui and Zhou, Yanmin and Ma, Jiajun and Jiang, Rong and Wang, Zhipeng and He, Bin},
	year = {2025},
}

@article{monguzzi_optimal_2025,
	title = {Optimal model-based path planning for the robotic manipulation of deformable linear objects},
	volume = {92},
	issn = {07365845},
	url = {https://linkinghub.elsevier.com/retrieve/pii/S0736584524001789},
	doi = {10.1016/j.rcim.2024.102891},
	language = {en},
	urldate = {2025-02-12},
	journal = {Robotics and Computer-Integrated Manufacturing},
	author = {Monguzzi, Andrea and Dotti, Tommaso and Fattorelli, Lorenzo and Zanchettin, Andrea Maria and Rocco, Paolo},
	month = apr,
	year = {2025},
}

@article{sanchez_robotic_2018,
	title = {Robotic manipulation and sensing of deformable objects in domestic and industrial applications: a survey},
	volume = {37},
	issn = {0278-3649, 1741-3176},
	shorttitle = {Robotic manipulation and sensing of deformable objects in domestic and industrial applications},
	url = {https://journals.sagepub.com/doi/10.1177/0278364918779698},
	doi = {10.1177/0278364918779698},
	language = {en},
	number = {7},
	urldate = {2025-02-06},
	journal = {The International Journal of Robotics Research},
	author = {Sanchez, Jose and Corrales, Juan-Antonio and Bouzgarrou, Belhassen-Chedli and Mezouar, Youcef},
	month = jun,
	year = {2018},
}

@article{yang_learning_2022,
	title = {Learning differentiable dynamics models for shape control of deformable linear objects},
	volume = {158},
	issn = {09218890},
	doi = {10.1016/j.robot.2022.104258},
	journal = {Robotics and Autonomous Systems},
	author = {Yang, Yuxuan and Stork, Johannes A. and Stoyanov, Todor},
	year = {2022},
}

@article{yu_global_2023,
	title = {Global {Model} {Learning} for {Large} {Deformation} {Control} of {Elastic} {Deformable} {Linear} {Objects}: {An} {Efficient} and {Adaptive} {Approach}},
	volume = {39},
	issn = {1552-3098},
	doi = {10.1109/TRO.2022.3200546},
	number = {1},
	journal = {IEEE Transactions on Robotics},
	author = {Yu, Mingrui and Lv, Kangchen and Zhong, Hanzhong and Song, Shiji and Li, Xiang},
	year = {2023},
}

@inproceedings{eberhard-2023-pink,
  title = {Pink Noise Is All You Need: Colored Noise Exploration in Deep Reinforcement Learning},
  author = {Eberhard, Onno and Hollenstein, Jakob and Pinneri, Cristina and Martius, Georg},
  booktitle = {Proceedings of the Eleventh International Conference on Learning Representations (ICLR 2023)},
  month = may,
  year = {2023},
  url = {https://openreview.net/forum?id=hQ9V5QN27eS}
}

\end{document}